\documentclass[letterpaper, 10 pt, conference]{ieeeconf}  

\IEEEoverridecommandlockouts                              

\usepackage{times}
\usepackage{microtype}
\usepackage{epsfig}
\usepackage[table,xcdraw]{xcolor}
\usepackage{caption}
\usepackage{float}
\usepackage{placeins}
\usepackage{color, colortbl}
\usepackage{stfloats}
\usepackage{tabularx}
\usepackage{xstring}
\usepackage{multirow}
\usepackage{xspace}
\usepackage{url}
\usepackage{xcolor}
\usepackage[hang,flushmargin]{footmisc}
\usepackage{array}
\usepackage{graphicx}
\usepackage{amsmath,amssymb}
\usepackage{multirow}
\usepackage{marvosym}
\usepackage{makecell}
\usepackage{bm}
\usepackage{float}
\usepackage{amsthm}

\usepackage{subfigure}
\usepackage{booktabs}
\usepackage{amsmath}
\usepackage{amssymb}
\usepackage{mathrsfs}
\usepackage[ruled,vlined,linesnumbered]{algorithm2e}
\usepackage{wrapfig}
\usepackage{hyperref}

\title{\LARGE \bf Non-Overlap-Aware Egocentric Pose Estimation for Collaborative Perception in Connected Autonomy
}

\author{Hong Huang$^{1}$, Dongkuan Xu$^{2}$, Hao Zhang$^{3}$, and Peng Gao$^{2}$
\thanks{$^{1}$Hong Huang works as a volunteer researcher with Prof. Peng Gao at NCSU. {Email: hhong$\_$@outlook.com}.
$^{2}$Dongkuan Xu and Peng Gao are with the Department of Computer Science, North Carolina State University, Raleigh, NC, USA. {Email:$\{$dxu27,pgao5$\}$@ncsu.edu}. $^{3}$Hao Zhang is with
Human-Centered Robotics Lab at University of Massachusetts Amherst, Amherst, MA, USA {Email: hao.zhang@umass.edu.}
}}

\begin{document}

\maketitle
\thispagestyle{empty}
\pagestyle{empty}

\begin{abstract}
Egocentric pose estimation is a fundamental capability for multi-robot collaborative perception in connected autonomy, such as connected autonomous vehicles. During multi-robot operations, a robot needs to know the relative pose between itself and its teammates with respect to its own coordinates. However, different robots usually observe completely different views that contains similar objects, which leads to wrong pose estimation. In addition, it is unrealistic to allow robots to share their raw observations to detect overlap due to the limited communication bandwidth constraint. In this paper, we introduce a novel method for \textit{Non-Overlap-Aware Egocentric Pose Estimation} (\textbf{NOPE}), which performs egocentric pose estimation in a multi-robot team while identifying the non-overlap views and satifying the communication bandwidth constraint. NOPE is built upon an unified hierarchical learning framework that integrates two levels of robot learning: (1) high-level deep graph matching for correspondence identification, which allows to identify if two views are overlapping or not, (2) low-level position-aware cross-attention graph learning for egocentric pose estimation. To evaluate NOPE, we conduct extensive experiments in both high-fidelity simulation and real-world scenarios. Experimental results have demonstrated that NOPE enables the novel capability for non-overlapping-aware egocentric pose estimation and achieves state-of-art performance compared with the existing methods. Our project page at \href{https://hongh0.github.io/NOPE/}{\emph{https://hongh0.github.io/NOPE}}.

\end{abstract}

\section{Introduction}

Multi-robot systems have attracted wide attention in recent decades due to their scalability~\cite{kuckling2024we}, parallelism~\cite{chen2025relative}, and reliability~\cite{park2017fault}. A fundamental capability in multi-robot systems is collaborative perception, which allows individual robots to share their own perception of the environments, thus leading to a shared situational awareness. It has lots of applications, such as connected autonomous driving~\cite{liu2023towards},~\cite{hu2024collaborative}, collaborative simultaneous localization and mapping (SLAM)~\cite{liu2024edge},~\cite{placed2023survey},~\cite{feng2024s3e}, and multi-robot search and rescue~\cite{chang2022lamp},~\cite{kashyap2023simulations}.

To enable efficient collaborative perception, it is essential to achieve accurate egocentric pose estimation that estimates the relative pose between a robot and its teammates with respect to its own coordinate. This allows each robot to determine the poses of its teammates, facilitating the aggregation of multi-robot perception. As shown in Figure~\ref{fig:motivation}, when connected autonomous vehicles meet at an intersection, the ego vehicle must first estimate the poses of its collaborators before merging their perceptions to improve situational awareness. This is particularly crucial in urban areas where GPS is unreliable or even unavailable. However, achieving accurate egocentric pose estimation presents two key challenges.  The first challenge arises from non-overlapping views, where different robots may observe totally different scenes while their different observations containing similar objects (e.g., traffic signs). This  can lead to wrong pose estimations.
The second challenge is limited communication bandwidth, which prevents vehicles from sharing raw observations to compare their observations to decide it they are overlapping or not.

\begin{figure}
\centering
\includegraphics[width=0.48\textwidth]{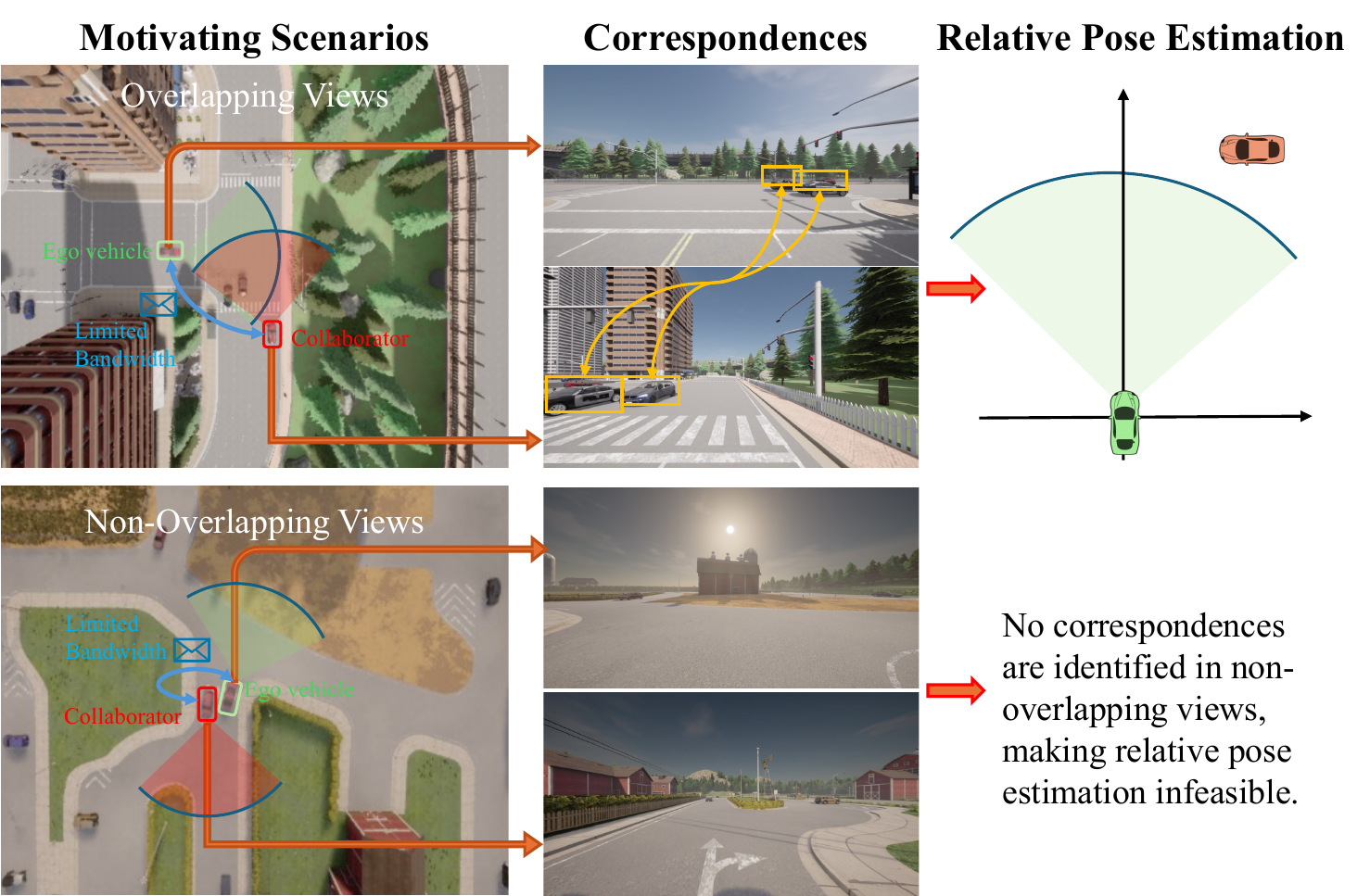}\label{fig:motivation}
\caption{A motivating scenario for egocentric pose estimation in connected autonomous driving. When two connected vehicles meet at an intersection, the ego vehicle must first estimate the pose of its teammate before merging its perception to enhance situational awareness. Meanwhile, it needs to address the challenges of limited communication bandwidth and non-overlapping views, where each vehicle observes a completely different perspective.
}\label{fig:motivation}
\vspace{-1em}
\end{figure}

Given the importance of egocentric pose estimation, a variety of methods have been studied. Previous techniques for multi-robot relative pose estimation often rely on SLAM, which assumes that robots share a global map \cite{miller2021any}  or utilize cross-robot loop closure to merge the local maps built by individual robots \cite{yin2021fusionvlad}.
However, merging local maps is both time- and bandwidth-intensive and typically struggles to handle dynamic changes in the environment.
Recently, vision-based methods have been developed using image registration through feature matching \cite{sun2021loftr,xie2024deepmatcher} or geometric alignment \cite{jiang2024se,wang2023posediffusion}.
However, according to the real-world setting, the maximum bandwidth designated for vehicle-to-everything (V2X) communication is around $7.2$ Mbps \cite{gallo2013short}, which is infeasible to share raw observations among robots.
In addition, there are non-overlapping views existing among multi-robot observations, which may contains similar objects. A unified framework to address all these challenges has not been well addressed yet.

To address these challenges, we propose a novel hierarchical learning approach called \textit{Non-Overlap-Aware Egocentric Pose Estimation} (\textbf{NOPE}), which performs egocentric pose estimation in a multi-robot team while identifying their non-overlap views. We represent each observation as a graph with nodes denoting objects associated with visual features extracted from the large vision model and edges denoting the spatial relationships of objects. Given the graph representations, our NOPE approach integrates two levels of robot learning into a hierarchical framework. The high-level NOPE performs correspondence identification (CoID) based on deep graph matching, which determines if two views are overlapped. The low level of NOPE utilizes a position-aware cross-attention network to capture the holistic context of observations for egocentric pose estimation.

The key contribution of the paper is the introduction of a novel approach to perform egocentric pose estimation while satisfying the communication bandwidth and identifying non-overlapping views. The specific novelties include:

\begin{itemize}
    \item This work presents one of the first learning solutions for multi-robot egocentric pose estimation with the awareness of non-overlapping views. It enables a novel multi-robot capability, allowing the ego robot to estimate the poses of its teammates while detecting non-overlapping views and satisfying the communication bandwidth constraint, thus enhancing situational awareness.
    \item We introduce a novel hierarchical learning approach that integrates a high-level deep graph matching network for non-overlap detection and a low-level position-aware cross-attention graph learning network for egocentric pose estimation.  Our approach achieves over {\bf 53$\%$} and {\bf 78.6$\%$} improvements on position and rotation estimations, as well as achieves over {\bf 96x} reduction in the shared data size that meets the realistic communication bandwidth constraint.  
\end{itemize}
\section{Related Work}\label{sec:related}
\subsection{Collaborative Perception}
Collaborative perception has gained significant attention in recent research. 
From an application perspective, collaborative object localization surpasses single-view localization by leveraging multi-robot observations to consistently identify the same objects~\cite{gao2022asynchronous,ji2017surfacenet,gao2021multi}. This technique enhances accuracy by fusing different viewpoints and addressing occlusions.  In addition, collaborative perception plays a crucial role in trajectory forecasting~\cite{zhu2021learning}, scene segmentation~\cite{liu2020who2com,liu2020when2com}, tracking, and object detection~\cite{robin2016multi}. These tasks benefit from associating multi-robot observations to build a richer, more robust environmental understanding. 
However, existing methods often assume that robots share overlapping observations, such as connected vehicles meeting at an intersection, which may not always be the case in more dynamic and unstructured environments.

From a solution perspective, collaborative perception is typically categorized into three approaches.
First, early fusion directly integrates raw sensor data from multiple robots before processing~\cite{arnold2020cooperative}. While this method retains the most information, it heavily depends on high-bandwidth communication, making it impractical in constrained network conditions. Second, intermediate fusion seeks a balance between information-sharing efficiency and computational cost by transmitting compressed feature representations instead of raw data. These methods include when2com~\cite{liu2020when2com}, who2com~\cite{liu2020who2com}, and where2com~\cite{hu2022where2comm}, which selectively share relevant features to optimize perception efficiency. However, these methods rely on coordinate transformations based on GPS or pre-existing maps to align observations, making them unreliable in GPS-denied environments or dynamically changing robot teams. Third, late fusion merges independent perception outputs from multiple robots, often using post-processing techniques like Non-Maximum Suppression (NMS)~\cite{forsyth2014object} and refined matching for pose consistency \cite{song2023cooperative}. While this approach is robust to noise and requires minimal bandwidth, it ignores most of the useful information in the raw data, which limits its adaptability to unknown observations.

\subsection{Multi-Robot Relative Pose Estimation}
The existing methods of multi-robot relative pose estimation can be divided into three categories, including GPS-based methods, SLAM-based methods, and vision-based methods. 
First, GPS-based methods rely on accurate GPS signals to provide coordinate transformations for estimating relative poses among robots, such as in UAVs~\cite{an2023array} and connected autonomous driving~\cite{gao2023deep,gao2023uncertainty}. However, GPS is often unreliable in highly dynamic environments and somtimes unavailable.
Second, SLAM-based methods assume that the entire robot team maintains a shared global map, with loop closure detection used to estimate egocentric poses relative to this map~\cite{miller2021any,gao2023visual,gao2020long}. However, global maps and reliable loop closures are not always available, especially in large-scale or dynamically changing environments, significantly limiting their applicability. 
Third, vision-based methods estimate relative poses by registering two observations (e.g., RGB images or point clouds) through feature matching~\cite{sun2021loftr,xie2024deepmatcher,sarlin2020superglue,zhang2024diffglue} or geometric alignment~\cite{jiang2024se,wang2023posediffusion, qin2022geometric}. While these methods can provide high accuracy, they need substantial communication costs, making them impractical for real-time robotic applications with limited resources.

Recently, foundational models have been widely used as strong priors to various applications due to the generalizability. Foundation visual models, such as DINO~\cite{oquab2023dinov2} and CLIP~\cite{radford2021learning}, have been extensively used to extract meaningful and generalizable representations for relative pose estimation \cite{blumenkamp2024covis}. Even though it achieves promising performance on egocentric pose estimation, it still cannot address scenarios where the views of two robots have no overlap.

\section{Approach}\label{sec:method}

\begin{figure*}
\centering
\includegraphics[width=0.9\linewidth]{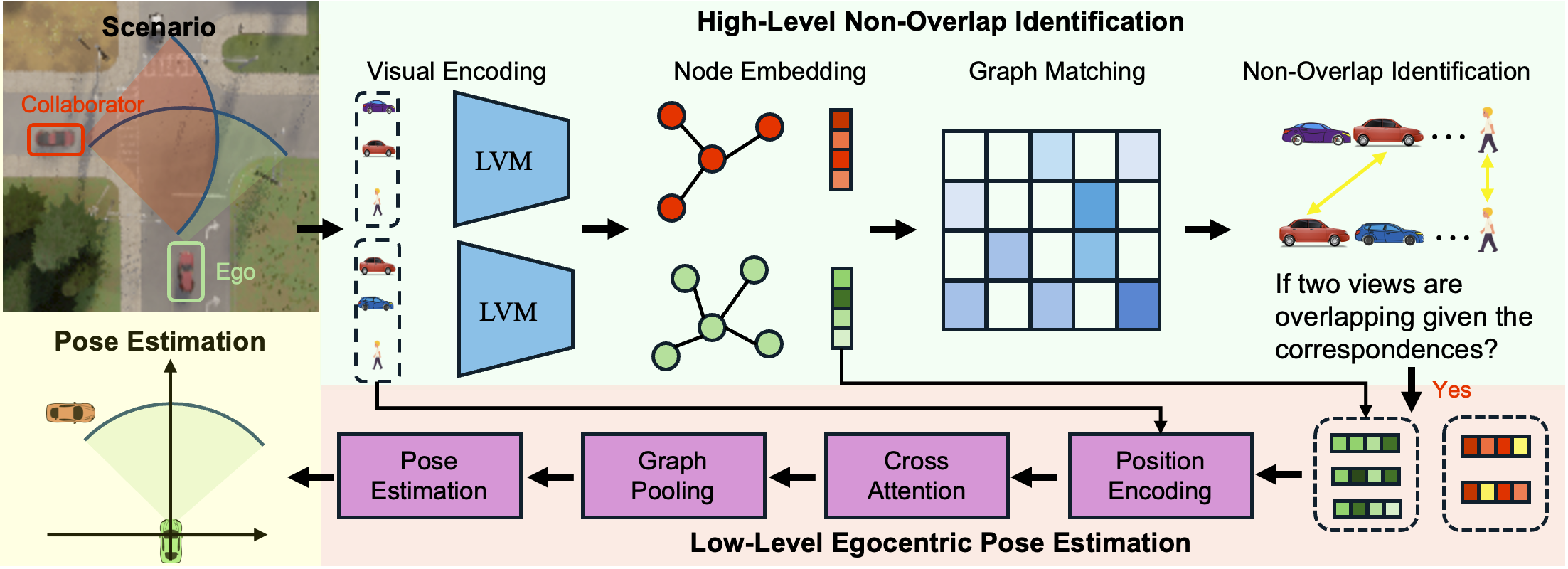}\label{fig:framework}
\caption{Overview of our NOPE framework. NOPE represents the observation of each robot as a graph. The high-level NOPE performs CoID based on LVM-based deep graph matching. The identified correspondences are used to detect the overlapping views. The low-level of NOPE utilizes a position-aware cross-attention graph learning network to perform pose estimation between the ego robot and its teammate robot.}
\vspace{-1em}
\end{figure*}

\subsection{Problem Definition}
For each robot, we represent its observations as a graph $\mathcal{G}(\mathcal{V}, \mathcal{F}, \mathcal{E})$.
The node set $\mathcal{V} = \{ v_1, v_2, \dots, v_n \}$ contains the objects observed by the robot, where $v_i \in \mathcal{V}$ denotes the 3D position of the $i$-th detected object (e.g., a vehicle or a pedestrian). Each object is associated with a visual feature vector, denoted as $\mathcal{F} = \{ \mathbf{f}_1, \mathbf{f}_2, \dots, \mathbf{f}_n \}$, where $\mathbf{f}_i$ represents the visual feature vector of the $i$-th object. The edge set $ \mathcal{E} \subseteq \mathcal{V} \times \mathcal{V} $ defines the spatial relationships of objects, where nodes $v_i$ and $v_j$ are connected if through the geometric rules of Delaunay triangulation, there exists a direct connection relationship between  $i$-th and $j$-th objects in the geometric space. Consequently, the adjacency matrix  $\mathbf{A}$  can be derived, where  $\mathbf{A}_{i,j} = \| v_i - v_j \|_2$  if nodes  $v_i$  and  $v_j$  are connected; otherwise,  $\mathbf{A}_{i,j} = 0$. 
The graph representation can significantly reduce the size of data shared among robots, thus satisfying the communication bandwidth constraints.

Given the graph representations $\mathcal{G}$ and $\mathcal{G^\prime}$ provided by a pair of robots,  we aim to address the problem of egocentric pose estimation with the identification of non-overlapping views:
\begin{enumerate}
    \item \textbf{Non-Overlap Detection}: The capability of a robot team identifying if their views are overlapped or not solely based on their visual perception. 
    \item \textbf{Egocentric Pose Estimation}: The capability of a robot to estimate the pose of its robot teammates with respect to its own egocentric coordinates. 
\end{enumerate}

\subsection{High-Level Correspondence Identification for Non-Overlap Detection}
Given the graph representations $\mathcal{G}$ and $\mathcal{G^\prime}$, we formulate non-overlap detection as a graph matching problem, which identifies the correspondences of objects in different views to determine if two views are overlapped or not.
First, we utilize large vision models (LVMs), such as DINO~\cite{oquab2023dinov2} or CLIP~\cite{radford2021learning}, to extract visual features of objects. Formally, it is defined as  
\begin{equation}
    \mathbf{f}_i = \Phi(I, v_i),
\end{equation}
where $\Phi(\cdot)$ denotes the LVM-based encoder, $I$ is the observation of a robot, and $v_i$ denotes the $i$-th object observed by the robot. 
Then, we employ a Transformer-based graph attention network  $\{\mathbf{h}_i\}^n=\Psi(\mathcal{F},\mathbf{A})$ to compute node embeddings, where $\mathbf{h}_i$ denotes the embedding of the i-th object, which does not just consider its own visual feature but also aggregating its neighbors. Formally, we compute the linear projection of the embedding of the $i$-th object as follows:
\begin{equation}\label{eq:2}
    \mathbf{q}_i^l = \mathbf{W}^{l,q} \mathbf{h}_i, \quad \mathbf{k}_i^l = \mathbf{W}^{l,k} \mathbf{h}_i, \quad \mathbf{v}_i^l = \mathbf{W}^{l,v} \mathbf{h}_i,
\end{equation}
where $\mathbf{q}_i^l$, $\mathbf{k}_i^l$, and $\mathbf{v}_i^l$ represent the query, key, and value vectors of the $i$-th object at layer $l$. The trainable weight matrices corresponding to these transformations are denoted as $\mathbf{W}^{l,q}$, $\mathbf{W}^{l,k}$, and $\mathbf{W}^{l,v}$. Notably, the initial input is defined as $\mathbf{h}^0_i = \mathbf{f}_i$.
The attention between pairs of nodes is computed as follows:
\begin{equation}\label{eq:3}
    \alpha_{i,j}^l = \frac{\exp((\mathbf{q}_i^l)^\top (\mathbf{k}_j^l+\mathbf{W}^{l,e}\mathbf{A}_{i,j}))}{\sqrt{d}\cdot\sum_{k \in \mathcal{N}(i)} \exp((\mathbf{q}_i^l)^\top (\mathbf{k}_k^l+\mathbf{W}^{l,e}\mathbf{A}_{i,k}))},
\end{equation}
where $\mathcal{N}(i)$ denotes the set of neighboring nodes of $v_i$, $\textbf{W}^l_e$ denotes a trainable weight matrix, and $d$ denotes the length of query vector. 
$\alpha^l_{i,j}$ denotes the attention from the $i$-th node to the $j$-th node, which is computed by comparing the query of the $i$-th node and its neighbors. The adjacent matrix $\mathbf{A}_{i,k}$ is added into the learning process to encode the spatial relationships of nodes. Then, SoftMax is used to normalize the attention. The final node embedding is computed as follows:
\begin{equation}
    \mathbf{h}_i^{l+1} = \text{LayerNorm} \left( \mathbf{W}^{l,h} \mathbf{h}_i^l + \Big\|_{m=1}^{M}\sum_{j \in \mathcal{N}(i)} \alpha_{i,j}^l \mathbf{v}_j^l \right),
\end{equation}
where the $\|$ is the concatenation operation for $M$ head attention, \text{LayerNorm} denotes layer normalization operation and $\mathbf{W}^{l,h}$ denotes a trainable weight matrix.
The final node embedding is computed by aggregating the central node embedding and its neighborhood node embeddings weighted by attention coefficients. Multi-head mechanism generates a richer representation of the embedding by capturing different embedding spaces and layer normalization standardizes the embeddings to improve training stability and convergence.

Given the node embeddings $\mathbf{h}_i^L$ and $\mathbf{h}_j^L$ from the final layer $L$, we compute pairwise correspondences between graphs $\mathcal{G}$ and $\mathcal{G}^\prime$. A similarity matrix $\mathbf{S}$ is computed as:
\begin{equation}
    \mathbf{S}_{i,j} = \mathbf{h}_i^L (\mathbf{h}_j^{L})^\top,
\end{equation}
where $ \mathbf{S} \in \mathbb{R}^{n \times m}$  represents the similarity matrix between two graphs containing $n$ and $m$ objects, respectively. To improve the robustness of CoID, a graph difference matrix $\mathbf{D}$ is to update the similarity matrix $\mathbf{S}$, which is defined as:
\begin{equation}
    \mathbf{D} = \left( \mathbf{S}^\top \Psi(\mathbf{J}, \mathbf{A}) - \Psi(\mathbf{S}^\top \mathbf{J}, \mathbf{A}^\prime) \right)^\top,
\end{equation}
where  $\mathbf{J} \in \mathbb{R}^{n \times r}$ is a random matrix. According to graph consensus theorem \cite{fey2020deep}, when the graphs $\mathcal{G}$ and $\mathcal{G}^\prime$  represent the same graph, then $\mathbf{S}^\top \Psi(\mathbf{J}, \mathbf{A}) = \Psi(\mathbf{S}^\top \mathbf{J}, \mathbf{S}^\top \mathbf{A} \mathbf{S}) = \Psi(\mathbf{S}^\top \mathbf{J}, \mathbf{A}^\prime)$, thus $\mathbf{D}_{i,j} = 0$. The larger the difference between two graphs, the large values in the difference matrix $\mathbf{D}_{i,j} = 0$. The updated similarity matrix is defined as:
\begin{equation}\label{eq:tau}
\hat{\mathbf{S}}_{i,j} = \epsilon\left(\left(\mathbf{S}_{i,j} + \mathbf{D}_{i,j}\right), \tau\right),
\end{equation}
where $ \hat{\mathbf{S}} \in \mathbb{R}^{n \times m}$  represents the final similarity matrix between two graphs. $\epsilon(\cdot)$ is an indicator function that outputs $1$  when $\mathbf{S}_{i,j} + \mathbf{D}_{i,j} \geq \tau$, otherwise $0$. $\tau$ denotes a threshold.
The final correspondences of objects are identified as follows:
\begin{align}\small\label{eq:y}
\mathbf{Y} &=\mathrm{argmax}_{\mathbf{Y}}\sum_{i=1}^{n}\sum_{j=1}^{m}\hat{\mathbf{S}}_{ij}\cdot\mathbf{Y}_{ij} \\
\mathrm{s.t.} & \quad \sum_{j=1}^{m}\textbf{Y}_{ij}\leq1,  \quad
\quad\sum_{i=1}^{n}\textbf{Y}_{ij}\leq1 \nonumber
\end{align}
where $\mathbf{Y} \in \{0,1\}^{n \times m}$ denotes the identified correspondences of objects observed by two robots, with $\mathbf{Y}_{i,j}=1$ denoting the $i$-th object observed by the ego robot and the $j$-th object observed by its teammate are the same.  The final correspondences are optimized by maximizing the overall similarity given the similarity matrix $\hat{\mathbf{S}}$. The constraint enforce that one object can at most have one corresponding object in the other observation, thus allowing to remove non-covisible objects that can only be observed by one robot. We use the Hungarian algorithm~\cite{kuhn1955hungarian} to solve this optimization problem.

Given the correspondence matrix $\mathbf{Y}$, we determine whether there exists an overlap between two observations. Specifically, if the sum of all elements in $\sum_{i,j}\mathbf{Y} =0$, it indicates that there is no overlap between two robots' views due to the lack of correspondences. If $\sum_{i,j}\mathbf{Y} \geq 1$, it implies an overlapping views between a pair of robots. As non-overlapping views significantly affect the pose estimation accuracy due to the lack of correlated contextual information, we only perform egocentric pose estimation when two observations are decided to be overlapping.
We train the high-level CoID network with the following loss function:
\begin{equation}
    \mathcal{L}^{high}=\frac{\sum_{i,j}(\hat{\mathbf{S}}_{i,j}-\mathbf{Y}^*_{i,j})}{n\cdot m},
\end{equation}
where $ \mathbf{Y}^* \in \mathbb{R}^{n \times m}$ denotes the ground truth correspondence matrix. If the $i$-th object in one observation and the $j$-th object in the other observation are the same, then: $\textbf{Y}^*_{i,j} = 1$, otherwise $0$. The correspondence is optimal when the loss is minimum.

\subsection{Low-Level Position-Aware Graph Learning for Egocentric Pose Estimation}
Once two observations are decided to be overlapping given the high-level CoID results, we design a low-level network based on position-based cross-attention mechanism to estimate the relative poses between the ego robot and its teammates.
To capture the holistic information of the observation for egocentric pose estimation, we compute the graph-level embeddings that captures the whole visual-spatial information of the observation as a single vector, meanwhile considering positional cues of node embeddings and the correlation between two observations, to improve the expressiveness of graph embeddings. 

First, we explicit encode the order of node embeddings as $\mathbf{P}_i = \mathbf{U}[i,:]$, where $\mathbf{P}_i$ denotes the position embedding of the $i$-th object in observations $\mathcal{G}$ and $\mathbf{U}$ denotes learnable embedding matrices. During training, the parameters of $\textbf{U}$ are optimized to capture meaningful positional information, allowing the model to learn an effective representation of position embeddings for each object in the sequence.

By incorporating position embeddings, our model gains the ability to discern the relative positions of objects in observations, thereby enhancing the accuracy of egocentric pose estimation.

Second, we compute a set of node embeddings $ \mathbf{H}$ of graph $\mathcal{G}$ by concatenating the node embeddings $\{\mathbf{h}_i\}^n$ and the further update it by combining the position embeddings: 
\begin{equation}
    \hat{\mathbf{H}}_i= \mathbf{H}_i + \mathbf{P}_i.
\end{equation}

Third, we use cross-attention mechanism to capture the attention of the relevant objects between two observations with partial overlap, thus improving the egocentric pose estimation. Formally, given the sets of node embeddings  $\hat{\mathbf{H}}$ and $\hat{\mathbf{H}^\prime}$ of graphs $\mathcal{G}$ and $\mathcal{G}^\prime$, we compute the cross attention to capture the correlations of nodes embeddings as follows:
\begin{equation}\small
    \text{CrossAtt}(\hat{\mathbf{H}}, \hat{\mathbf{H}}^\prime) = \text{SoftMax} \left( \frac{\hat{\mathbf{H}} \mathbf{W}^Q (\hat{\mathbf{H}}^\prime \mathbf{W}^K)^\top}{\sqrt{d}}\right) ,
\end{equation}
where  $\mathbf{W}^Q$, $\mathbf{W}^K$ and $\mathbf{W}^V$ denotes the trainable weight matrices. By comparing the similarity between the sets of node embeddings $\hat{\mathbf{H}}$ and $\hat{\mathbf{H}}^\prime$, the cross attention between two observations  $\mathcal{G}$ and $\mathcal{G}^\prime$ is computed through a SoftMax function. Then we update the set of node embedding as follows:
\begin{equation}\small
    \mathbf{H}_\text{out}=\text{LayerNorm}(\hat{\mathbf{H}} +\text{MLP}(\Big\|_{m=1}^{M}\text{CrossAttn}(\hat{\mathbf{H}}, \hat{\mathbf{H}}^\prime)_m\cdot\hat{\mathbf{H}} \mathbf{W}^V),
\end{equation}
We use multi-head mechanism followed by a MLP to update the set of node embeddings based on cross attention. Then the updated graph embedding and the original graph embedding $\hat{\mathbf{H}}$ are added up and pass through a layer normalization to generate $\mathbf{H}_\text{out}$.

Finally, we employ an attention gate aggregation operation to compute the graph-level embeddings. Specifically, we apply self-attention to $\textbf{H}_\text{out}$, which captures the weighted relationships between nodes by considering the interactions and dependencies within the node embeddings.
\begin{equation}\label{eq:aggr1}
    \mathbf{H}_\text{weight}=\text{SoftMax}(\frac{\mathbf{H}_{\text{out}}\textbf{W}^Q(\mathbf{H}_{\text{out}}\textbf{W}^K)^\top}{\sqrt{d}})\mathbf{H}_{\text{out}}\textbf{W}^V,
\end{equation}

Then, we pass $\mathbf{H}_\text{weight}$ through a multi-layer perceptron (MLP) followed by a SoftMax function to obtain the attention gate scores $\mathbf{g}$.
\begin{equation}\label{eq:aggr2}
    \mathbf{g}=\text{SoftMax}(\text{MLP}(\mathbf{H}_\text{weight})),
\end{equation}
where  $\mathbf{g}$ denotes the attention gate score, which captures the importance of each node for graph embedding. The graph embedding is computed by pooling node embeddings $\mathbf{H}_{\text{out},i}$ weighted by the attention gate scores $\mathbf{g}_i$.
\begin{equation}\label{eq:aggr3}
    \mathbf{h}_{\text{pooled}} =  \sum_{i=1}^n \mathbf{g}_i\cdot\mathbf{H}_{\text{out},i},
\end{equation}
where $\mathbf{h}_{\text{pooled}}$ denotes the graph embedding of the graph $\mathcal{G}$, which captures all visual-spatial cues of nodes while considering the  correlation of observations provided by the ego robot and its collaborator. Given the graph embedding, we estimate the egocentric pose as follows:
\begin{equation}
    (\hat{\mathbf{p}}, \hat{\mathbf{R}})=\text{MLP}(\mathbf{h}_{\text{pooled}}).
\end{equation}
where  $\hat{\mathbf{p}}$ and $ \hat{\mathbf{R}}$ denote the position and rotation of the collaborator providing observation $\mathcal{G^\prime}$. The egocentric pose of an ego vehicle's collaborator is estimated from the position-aware cross-attention graph embedding $\mathbf{h}_{\text{pooled}}$ followed by a MLP.
The loss function to train the low-level egocentric pose estimation is defined as follows:
\begin{equation}\small
    \mathcal{L}^{Low} =\overbrace{\| \hat{\mathbf{p}} - \mathbf{p} \|_2^2}^{\mathcal{L}_{pos}} + \overbrace{2\cdot \|\hat{\mathbf{R}} - \mathbf{R} \|_2^2\cdot(4-\|\hat{\mathbf{R}} - \mathbf{R} \|_2^2)}^{\mathcal{L}_{rot}}.
\end{equation}

The first term denotes the position loss  which is compute by minimizing the Euclidean distance between the predicted position $\hat{\mathbf{p}}$ and the ground truth position $\mathbf{p}$. The second term denotes the loss of rotation estimation with respect to the ego robot coordinates. It is based on the  chordal squared loss~\cite{blumenkamp2024covis} to measure the difference between the quaternion-based rotation estimation $\hat{\mathbf{R}}$ and the ground truth $\mathbf{R}$.

\section{Experiments}\label{sec:experiment}
\subsection{Experimental Setup}
We conducted experimental evaluation in both high-fidelity simulation and the real world. In the simulation, we utilize both CARLA~\cite{dosovitskiy2017carla} and SUMO~\cite{krajzewicz2002sumo} to create five connected autonomous driving (CAD) scenarios. In each scenario, a pair of connected vehicles are deployed. The behaviors of vehicles and pedestrians were controlled by SUMO in accordance with real-world rules, such as stopping at red lights and yielding to pedestrians. For each vehicle, it is equipped with a front-facing RGB-D camera  and a Global Navigation Satellite System (GNSS) sensor.  In the real-world application, we utilize the multi-modal autonomous driving dataset MARS~\cite{li2024multiagent}, which was collected by a fleet of autonomous vehicles operating within a specific geographic area. Each vehicle follows its own route, with different vehicles potentially appearing in nearby locations. Each vehicle was equipped with one LiDAR, three narrow-angle RGB cameras, three wide-angle RGB fisheye cameras, one IMU, and one GPS. All sensor data were sampled at 10Hz to ensure synchronization.

In the simulation, we collect a total of 30,277 data instances, of which 27,247 were used for training and 3,030 for testing. Each data instance includes a pair of RGB-D images captured from different perspectives by two connected vehicles.
The ground truth of object correspondences is provided by the CARLA simulation and the ground truth of positions and orientations of connected vehicles is provided by the GNSS sensor.
In the real-world application, we select 201 data instances. Each data instance consists of a pair of RGB images captured from different perspectives by two connected vehicles, along with the positions and orientations of the vehicles provided by GPS.

For graph construction, we use YOLOv5~\cite{jocher2022ultralytics} to detect objects in each vehicle's view and extract visual features as node representations. Delaunay triangulation generates edges and DepthAnythingV2~\cite{depth_anything_v2} estimates depth information, which is used to compute edge attributes based on object locations. In simulation CAD, vehicle positions and rotations are represented in XYZ-pitch-roll-yaw. In real-world CAD, they are in XYZ format with quaternion rotation.

In the specific implementation details, we implemented the Transformer-based graph attention etwork $\Psi$ using PyTorch and PyG~\cite{Fey/Lenssen/2019}. In this network, we set the number of network layers as $L = 2$, with the number of heads as $\text{heads} = 4$, and all dimensions $d = 256$. Additionally, the edge feature dimension is set to $\text{dim} = 1$. After each attention layer, we applied a dropout with a probability of 0.5. For the MLP with two linear layers, each layer has a dropout probability of 0.2. In the position-aware cross-attention network, the number of network layers is $L = 4$, with $\text{heads} = 4$, and $d = 256$. The attention gate aggregation network has $L = 1$ layers, with $\text{heads} = 4$, and $d = 256$. In all experiments, we used Adam as the optimizer~\cite{kingma2014adam}, with a learning rate of 0.001. We ran the training for 150 epochs.

We implement a baseline method $\textbf{NOPE}_\text{high}$ that just use the high-level network to evaluate the CoID performance. In addition, compare our method with six existing methods, including: 1) \textbf{GCN-GM}~\cite{fey2018splinecnn} aggregating visual-spatial information of objects and their neighborhoods via spline kernels for graph matching, 2) \textbf{DGMC}~\cite{fey2020deep} identifying initial correspondences based on visual similarity and graph matching consensus for CoID, 3) \textbf{BDGM}~\cite{gao2021bayesian} performing deep graph matching under a Bayesian framework to remove invisible objects given the quantified correspondence uncertainty, 4) \textbf{DMGM}~\cite{gao2023deep} considering visual-spatial cues, matching consensus and uncertainty of correspondences for CoID, 5) \textbf{SuperGlue}~\cite{sarlin2020superglue} optimizing feature matching as a differentiable optimal transport problem, which leverages Transformer-based graph neural networks to estimate relative poses between two graphs,
6) \textbf{CoViS-Net}~\cite{blumenkamp2024covis} is a foundation model for egocentric pose prediction, which utilizes DINOV2 as encoder on individual robots and generates bird-eye-view map to encode poses of a robot team.
As \textbf{GCN-GM}, \textbf{DGMC}, \textbf{BDGM} and \textbf{DMGM} do not have the capability of estimating poses, we use learning-free pose estimation method RANSAC~\cite{fischler1981random} and essential matrix decomposition to estimate the poses given the correspondences identified.

We use the following metrics to evaluate our NOPE, including

1) Precision is defined as the ratio of correctly retrieved correspondences to the  retrieved correspondences, 2) Recall is defined as the ratio of correctly retrieved correspondences to the  ground truth correspondences, 3) F1-Score, which evaluates the overall performance of the CoID method, is calculated as F1=$\frac{2\times\text{Precision}\times\text{Recall}}{\text{Precision}+\text{Recall}}$, 4) Position Error (\textbf{PE}) measures the Euclidean distance between the estimated and the ground truth position, 5) Rotation Error (\textbf{RE}) is defined similarly to the rotation loss, where the quaternion and the ground truth quaternion is used as a surrogate for the geodesic distance~\cite{blumenkamp2024covis}, 6) Packet Size ((\textbf{PS})) refers to the size of the data transmission packets shared between connected vehicles, to evaluate communication efficiency, 7) Non-Overlapping Detection Accuracy (\textbf{NDA})) is defined as the ratio of correctly detected non-overlapping observation pairs to the total number of observation pairs, to evaluate accuracy of non-overlapping identification.

\subsection{Results over Connected Autonomous Driving Simulations}

\begin{table}[t]
    \centering
    \tabcolsep=0.25cm
    \caption{Quantitative results of egocentric pose estimation in both simulation and the real world based on metrics of position error (PE), rotation error (RE) and package size (PS). The improvements is computed w.r.t CoViS-Net~\cite{blumenkamp2024covis}}\label{tab:pose}
    \begin{tabular}{ccccccccc}
    \toprule
        \multirow{2}{*}{Method}  
        & \multicolumn{2}{c}{CAD Simulation } 
        & \multicolumn{2}{c}{Real World } 
        & \multirow{2}{*}{PS $\downarrow$} \\
        \cmidrule(lr){2-3} \cmidrule(lr){4-5}
        & PE $\downarrow$ & RE $\downarrow$ 
        & PE $\downarrow$ & RE $\downarrow$ \\ 
        \midrule
        GCN-GM~\cite{fey2018splinecnn} 
        & 20.32 & 3.641 & 19.77 & 5.541 & 36.7KB \\ 
        DGMC~\cite{fey2020deep}  
        & 20.17 & 3.120 & 18.15 & 4.620 & 36.2kB \\ 
        BDGM~\cite{gao2021bayesian} 
        & 19.35 & 2.944 & 17.81 & 4.357 & 36.5KB \\ 
        DMGM~\cite{gao2023deep} 
        & 18.69 & 2.256 & 17.28 & 4.621 & 35.1KB \\ 
        SuperGlue~\cite{sarlin2020superglue} 
        & 18.80 & 2.406 & 15.03 & 3.679 & 2.3MB \\ 
        CoViS-Net~\cite{blumenkamp2024covis} 
        & 13.92 & 2.037 & 14.52 & 3.120 & 0.75MB \\ 
        \midrule  
        \textbf{NOPE} 
        & \textbf{6.42} & \textbf{0.435} & \textbf{13.63} & \textbf{2.226} & \textbf{27.0KB} \\ 
        Improvements(\%)  & 53.87 & 78.64 & 6.32 & 28.65 & 96.48 \\
    \bottomrule
    \end{tabular}
    \vspace{-1em}
\end{table}

The CAD simulation includes a lots of challenges to perform egocentric pose estimation, including highly dynamic street objects (e.g., pedestrians and vehicles) with ambiguous visual appearance caused by occlusion and long-distance observation, a large number of  non-covisible objects, limited communication bandwidth, as well as non-overlapping views between pairs of connected vehicles. We run our approach on a Linux machine with an i7 32-core CPU and 16G memory. The average execution time is around $20$Hz.

\begin{figure*}[tb]
\centering
\vspace{-1em}
\subfigure[Simulation CAD]{\includegraphics[height=6.7cm]{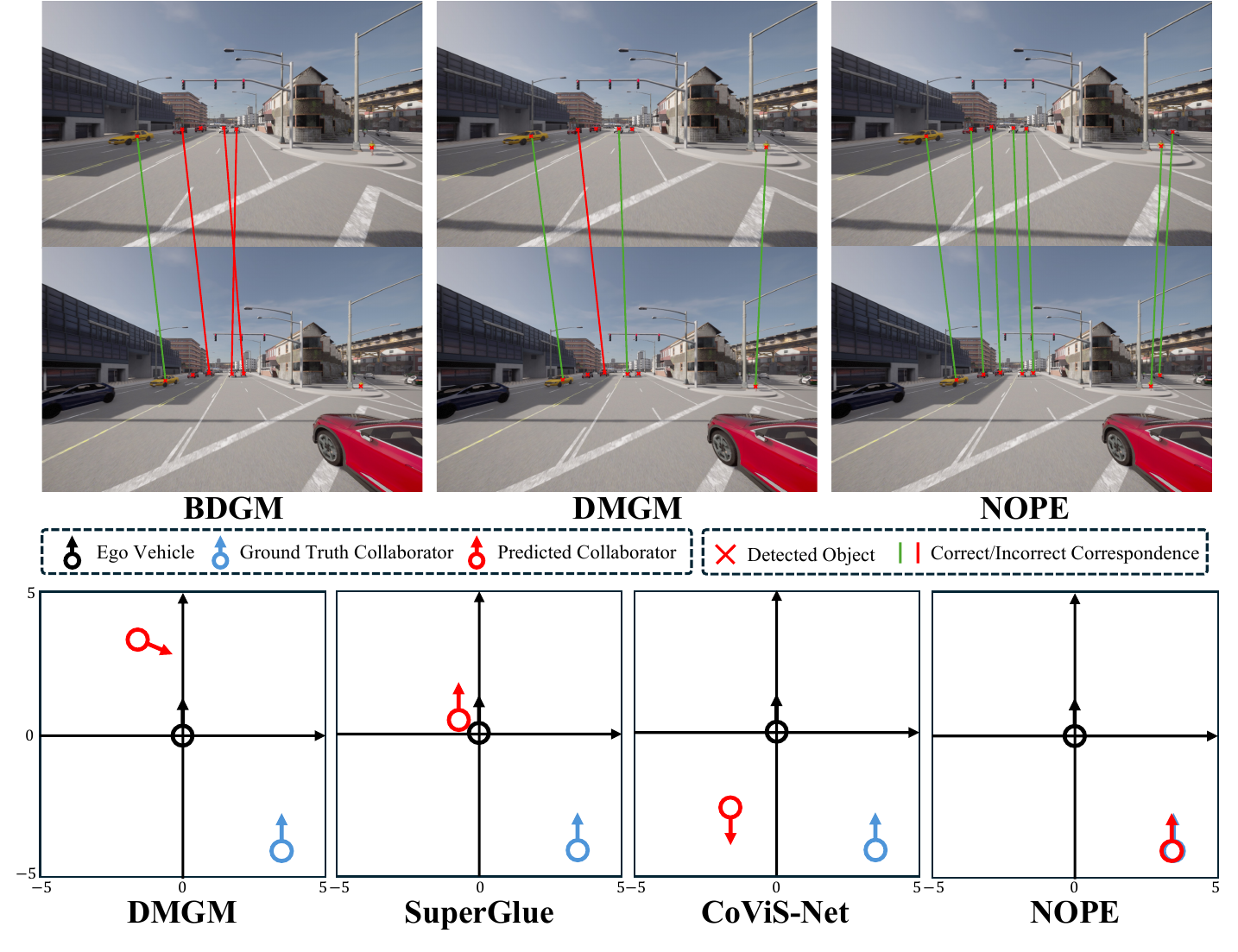} \label{fig:vis1}}
\subfigure[Real-world CAD]{\includegraphics[height=6.7cm]{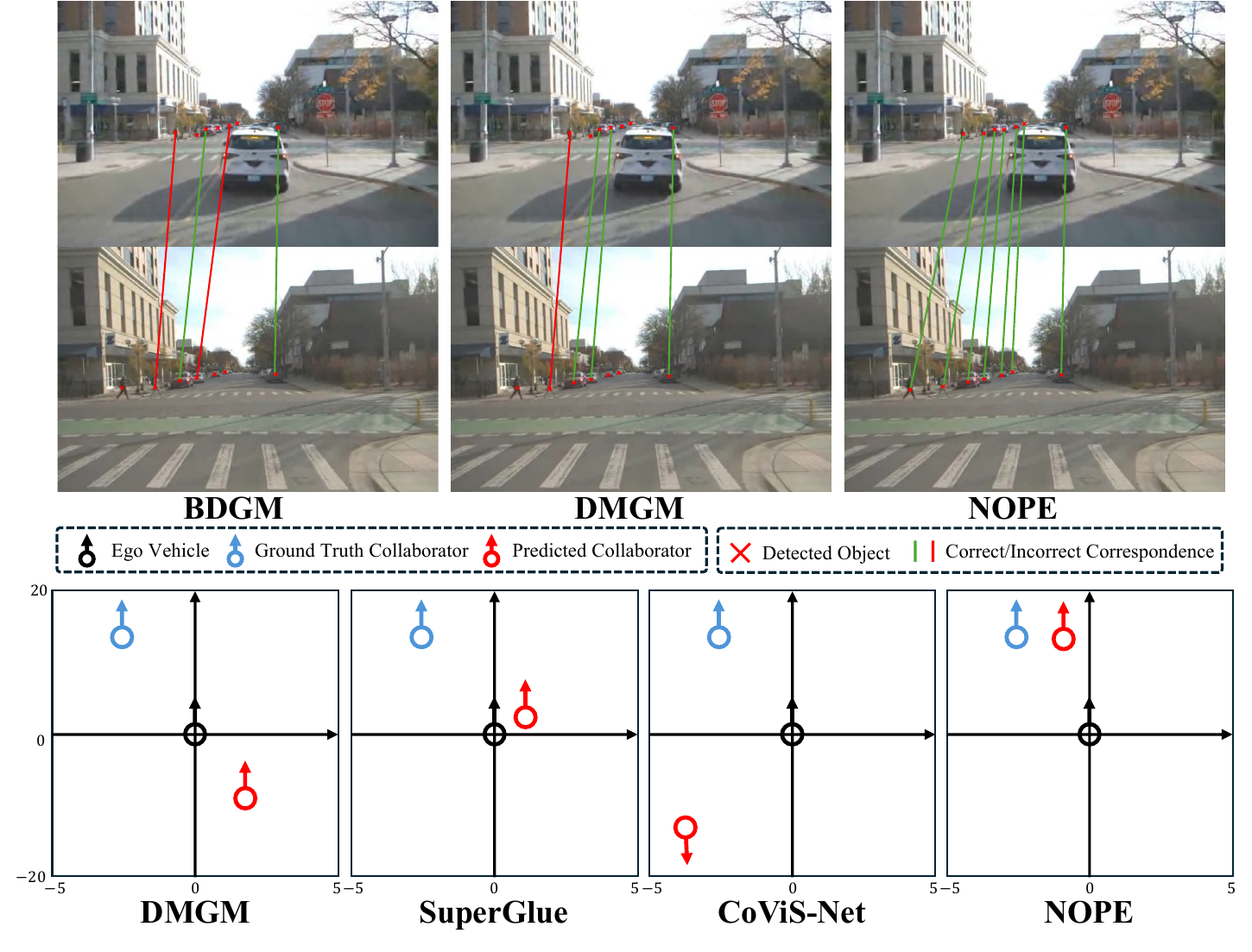}\label{fig:vis2}}
\caption{Qualitative results on CoID and egocentric pose estimation from both simulation and real-world scenarios. The first row illustrates identified correspondences between the ego robot and its collaborator's observations. The second row compares the estimated and ground truth poses of the collaborator in the ego vehicle's coordinate frame.
}	\label{fig:qual}
\vspace{-1em}
\end{figure*}

As shown in Figure~\ref{fig:vis1}, for CoID, we can clearly see that our NOPE outperforms existing methods BDGM and DMGM.
This is because of the integration of LVMs and addressing non-covisible objects in NOPE. For pose estimation,  the keypoint-based methods, SuperGlue and CoVisNet, can not well address the ambiguity in visual appearance of objects caused by long-distance observation and low resolution, which leads to poor pose estimations.  Moreover, as none of these existing method can address non-overlapping views,
NOPE achieves the best performance of egocentric pose estimation, which indicates the importance of addressing visual ambiguity, non-covisiblity and non-overlapping views for egocentric pose estimation in collaborative perception.

The quantitative results are shown in Table~\ref{tab:pose}. We observe that GCN-GM, DGMC, BDGM, and DMGM exhibit large pose errors. This is primarily because they focus on  correspondences of objects and rely on RANSAC for pose estimation, which requires a large number of correct correspondences of objects.
SuperGlue and CoVisNet achieve better performance by learning keypoint-based matching and pose estimation. However, these methods are highly sensitive to non-overlapping views with similar visual features.
NOPE outperforms CoVisNet, the second-best method, by 53.9\% in position estimation and 78.6\% in rotation estimation, while requiring only 1/96 of the data size for sharing.

\begin{table}[ht]
    \centering
    \tabcolsep=0.2cm

    \caption{Quantitative results of CoID in the simulation CAD based on metrics of precision, recall and non-overlapping detection accuracy (NDA). }\label{tab:coid}
    \begin{tabular}{ccccc}
    \toprule
        Method & Precision $\uparrow$ & Recall $\uparrow$ & F1-score  $\uparrow$ & NDA $\uparrow$ \\ 
        \midrule
        GCN-GM~\cite{fey2018splinecnn} & 0.5001 & 0.6391 & 0.5611 & 0.6539  \\ 
        DGMC~\cite{fey2020deep} & 0.4736 & 0.6425 & 0.5453 & 0.6857\\ 
        BDGM~\cite{gao2021bayesian} & 0.6817 & 0.6097 & 0.6437 & 0.7239\\ 
        DMGM~\cite{gao2023deep}  & 0.7859 & 0.8278 & 0.8063 & 0.7561 \\ 
        \midrule
        $\textbf{NOPE}_\text{high}$ & \textbf{0.8224} & \textbf{0.8429} & \textbf{0.8325} & \textbf{0.8039}\\
    \bottomrule
    \end{tabular}
\end{table}

\noindent
We further evaluate NOPE's high-level CoID for non-overlapping view detection in CAD simulation. Table~\ref{tab:coid} indicate that  GCN-GM and DGMC suffer from low recall due to their inability to handle non-covisible objects. BDGM prioritizes precision at the cost of recall based on the thresholding on the correspondence uncertainty but it uses hand-craft features for graph matching. By integrating LVMs and address non-covisible objects,
NOPE surpasses all existing methods on all metrics with the improvements of 4.6\%, 1.8\%, and 3.2\% on precision, recall and F1 score respectively. 
According to the metrics of NDA, we can see that our method is able to  identify over 80\% non-overlap views, which outperforms all the existing methods, which indicates the importance of identifying correct correspondences for the detection of non-overlapping views.

\subsection{Results over Real-World Connected Autonomous Driving}

The real-world connected autonomous driving scenario covers the challenges like low-resolution observations, limited communication bandwidth, non-covisible objects and non-overlapping views between connected vehicles. In addition, we do not fine tune NOPE with the real-world data and directly use the model learned from the CAD simulation to evaluate the generalizability of NOPE.

The qualitative results shown in Figure~\ref{fig:vis2} illustrates that other methods shows significant errors in  complex and noisy real-world scenarios in terms of both CoID or egocentric pose estimation.
In addition, NOPE still outperforms the existing methods 
without the needs of fine tuning, which indicates its generalizability to real-world applications.

Table~\ref{tab:pose} provides the quantitative results of egocentric pose estimation. We can observe that NOPE continuously maintains a low pose error compared with the second best method CoViS-Net. Although its rotation error is $2.226$ which slightly worse in the real-world scenario compared to its performance in the simulation, likely due to sensitivity to dynamic lighting, it still outperforms other methods. Furthermore, compared to feature-matching-based methods like SuperGlue and CoViS-Net, NOPE achieves the best results under realistic communication bandwidth constraints.

\subsection{Discussion}

\begin{figure}
\centering
\includegraphics[width=0.485\textwidth]{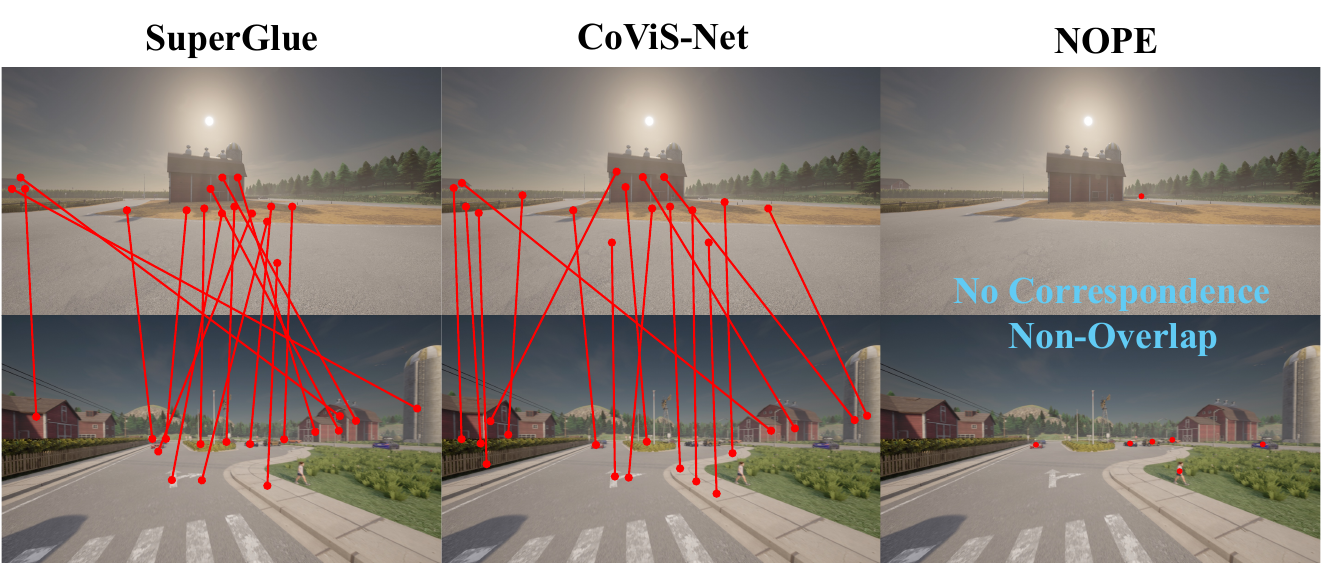}
\caption{Comparisons of CoID for non-overlap detection.}\label{fig:non-overlap}
\vspace{-1.5em}
\end{figure}

\textbf{Non-Overlapping Detection:}
Figure~\ref{fig:non-overlap} demonstrates that NOPE can detect the non-overlapping views by making decisions on the identified correspondences. When there is no identified correspondences of objects, NOPE determinate that two observations are non-overlapping, thus will not estimate the poses between two views. Notably, SuperGlue and CoViS-Net  rely on the keypoint feature matching, which can not work well with non-overlapping observations, particularly in different observations with similar visual context (e.g., traffic signs or buildings), which emphasizes the importance of CoID for non-overlapping view identification.

\begin{wrapfigure}{r}{0.2\textwidth}
    \centering
    \includegraphics[width=0.18\textwidth]{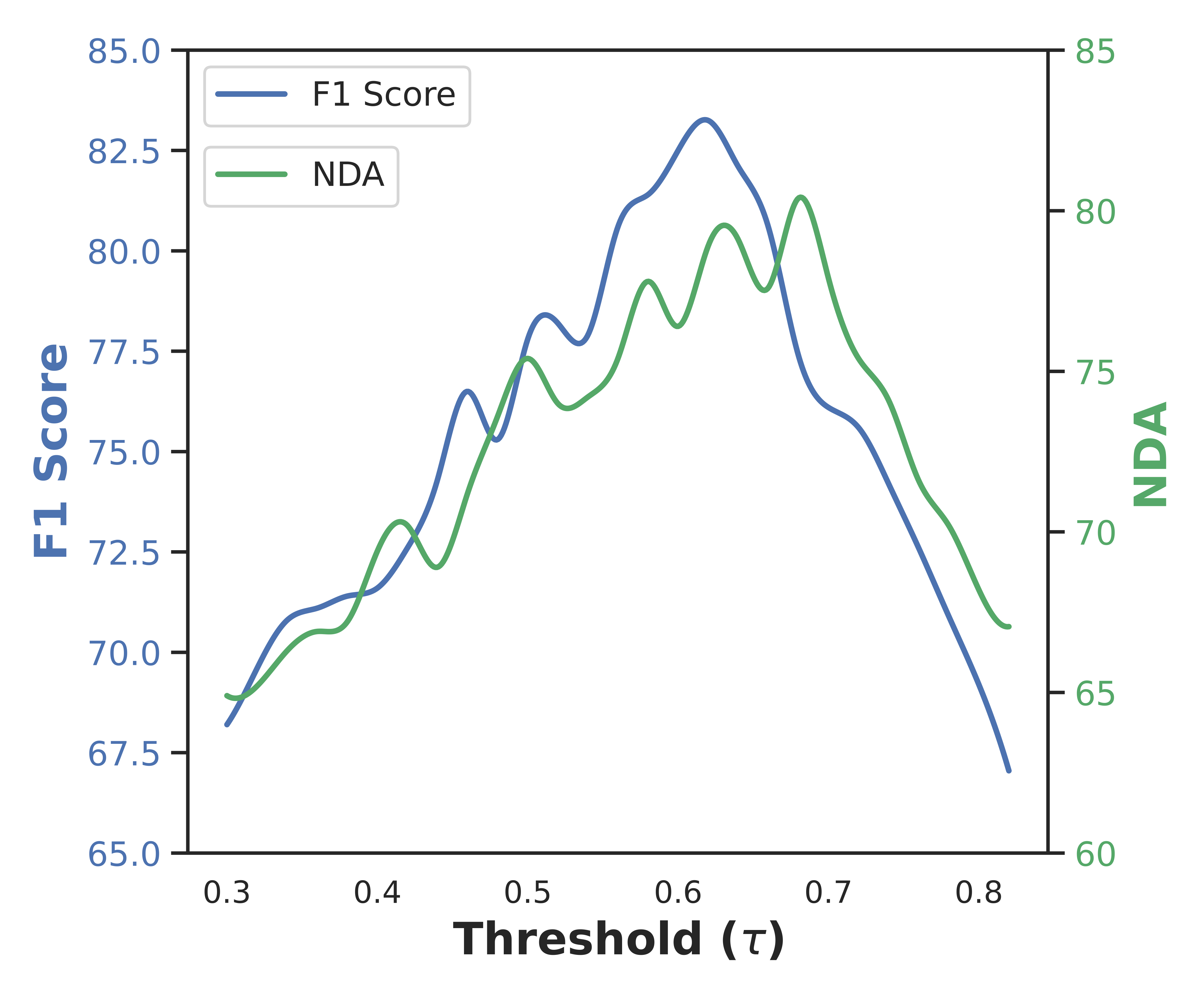}
    \vspace{-6pt}
    \caption{Analysis of $\tau$.}
    \label{fig:threshold}
    \vspace{-10pt}
\end{wrapfigure}
\textbf{Hyperparameter Analysis}: Figure~\ref{fig:threshold} shows the performance of our high-level CoID for non-overlapping detection with the variance of threshold $\tau$, defined in Eq. (\ref{eq:tau}). We can see that the highest non-overlap detection accuracy is achieved when $\tau$ is in the range of $[0.6, 0.7]$ with small fluctuation. Similarly, the F1 score reach to the highest when $\tau$ is in the range of $[0.6, 0.7]$, which indicates
positively correlation between the non-overlapping view detection and CoID.

\section{Conclusion}\label{sec:conclusion}

In this paper, we propose NOPE as a novel method to enable non-overlap-aware egocentric pose estimation for collaborative perception in multi-robot systems. NOPE integrates high-level deep graph matching to detect the overlap between two observations based on the identified correspondences, and low-level position-aware cross-attention network performs egocentric pose estimation. We conduct extensive experiments to evaluate NOPE in both high-fidelity simulation and real-world scenarios. The results demonstrate that NOPE enables new capability of non-overlap-aware egocentric pose estimation and significantly outperforms existing methods on bandwidth cost, non-overlap detection and egocentric pose estimation.

Our approach has several limitations that open avenues for future research.
First, NOPE cannot estimate egocentric poses when observations are completely non-overlapping. A possible solution is to integrate Bayesian filters to estimate relative poses, using NOPE’s outputs as corrections to refine these estimates.
Second, NOPE does not ensure global pose consistency for teams larger than two. Future work could explore distributed consensus algorithms to enable large robot teams to collaboratively achieve consistent pose estimation.










\bibliographystyle{IEEEtran}
\bibliography{ref}

\end{document}